\PassOptionsToPackage{table}{xcolor}
\documentclass[11pt]{article}

\usepackage[final]{acl}

\usepackage{times}
\usepackage{latexsym}
\usepackage{enumitem}

\usepackage[T1]{fontenc}

\usepackage[utf8]{inputenc}

\usepackage{microtype}

\usepackage{inconsolata}

\usepackage{graphicx}
\usepackage{comment}
\usepackage{multirow}
\usepackage{xcolor}
\usepackage{booktabs}

\title{Example-Guided Prompting for Document-Level Text Simplification}

\author{Ariel Perstin\\
  SCE, Beer Sheva, Israel \\
  \texttt{arielperstin10@gmail.com} \\\And
  Ilan Shtilman\\
 SCE, Beer Sheva, Israel \\
  \texttt{shtilmanilan@gmail.com} \\\AND
  Michael Färber\\
  ScaDS.AI, TUD\\ Dresden, Germany \\
  \texttt{michael.faerber@tu-dresden.de} \\\And Marina Litvak\\
  SCE, Beer Sheva, Israel \\
  ScaDS.AI, TUD\\
  Dresden, Germany \\
  \texttt{marinal@sce.ac.il}}

\begin{document}
\maketitle
\begin{abstract}
Document-level text simplification requires large language models (LLMs) to rewrite complex documents while preserving meaning, readability, and discourse coherence. Although prompt-based LLMs have shown promising performance, they often produce inconsistent simplifications because textual instructions alone provide limited guidance for complex document-level transformations. We investigate whether retrieved document-simplification examples can improve document-level generation by augmenting prompts with examples selected from a parallel simplification corpus. This example-guided prompting approach enables LLMs to exploit relevant simplification patterns without task-specific fine-tuning. Experiments on the OneStopEnglish corpus using multiple state-of-the-art LLMs show that incorporating retrieved examples consistently improves simplification quality over prompt-only generation and achieves competitive or superior performance compared with representative supervised (T5) and planning-based (PlanSimp) document simplification systems.  
Furthermore, we find that the benefits of example-guided prompting vary across LLMs, suggesting that effective use of retrieved examples depends on a model's ability to integrate contextual information during generation.
\end{abstract}

\section{Introduction}
Large language models (LLMs) have become an effective approach for text simplification, enabling high-quality rewriting through instruction following and in-context learning without task-specific training \cite{brown2020language, liu2023pre, cohen2026simplify}. While recent research has primarily focused on sentence-level simplification, many practical applications, such as educational resources, news articles, and technical documents, require simplifying entire documents while preserving semantic fidelity, discourse coherence, and an appropriate level of readability. Compared with sentence-level rewriting, document-level simplification requires coordinated transformations across multiple sentences while maintaining a consistent simplification style throughout the document.

Prompt-based LLMs provide a flexible solution, yet textual instructions alone often provide insufficient guidance for complex document-level transformations, leading to inconsistent simplification quality~\cite{cohen2026simplify, maddela2023lens, alfear2024meta}. This raises the question of whether retrieved examples can provide more effective guidance than textual instructions alone. 

Retrieval-Augmented Generation (RAG) has successfully improved many generation tasks by incorporating relevant external context during inference~\cite{lewis2020retrieval}. Unlike knowledge-intensive applications, document-level simplification requires guidance on how to transform text rather than additional factual information. We therefore investigate retrieval of document–simplification examples. 


To this end, we propose \textbf{Example-Guided Prompting (EGP)}, an inference-time approach that augments a strong prompt with retrieved examples selected from a parallel simplification corpus. Instead of relying solely on textual instructions, the model is exposed to examples illustrating how similar documents have previously been simplified, enabling it to exploit relevant simplification patterns without additional training or parameter updates.

We evaluate EGP on the OneStopEnglish corpus \cite{vajjala2018onestopenglish} using multiple state-of-the-art LLMs. Building on a strong prompt formulation, we investigate whether retrieved examples improve document-level simplification compared with prompt-only generation. We further compare EGP with two representative document simplification baselines: T5~\cite{raffel2020exploring} and PlanSimp, a document-level planning approach proposed by \cite{cripwell2023document}, using automatic measures of simplification quality~\cite{xu2016optimizing}, semantic preservation~\cite{maddela2023lens,zhang2019bertscore}, and readability \cite{flesch1948new,kincaid1975}.

Our experiments demonstrate that EGP consistently improves prompt-only generation, achieves competitive or superior performance compared with representative state-of-the-art simplification systems, and reveals that the effectiveness of example-guided prompting depends on the underlying LLM. 

Our contributions are summarized as follows: (1) We propose Example-Guided Prompting (EGP), an inference-time approach that augments LLM prompts with retrieved document–simplification examples for document-level text simplification\footnote{Our code and data are available at \url{https://github.com/simplify120/RAG-simp}. The link is anonymized to meet the double-blind reviewing policy.}. 
(2) We demonstrate that EGP consistently improves prompt-only generation across several state-of-the-art LLMs and achieves competitive or superior performance compared with representative supervised and planning-based document simplification systems. 
(3) We provide an empirical analysis of when example-guided prompting is most effective, showing that its benefits depend on the underlying LLM and involve a trade-off between semantic preservation and readability. 

\section{Background}

Text simplification aims to transform complex text into simpler versions while preserving meaning and improving accessibility for diverse audiences \cite{siddharthan2006syntactic, xu2015problems}. Early approaches relied on lexical substitution, syntactic rewriting, statistical machine translation~\cite{siddharthan2006syntactic, pavlick2016simple, xu2016optimizing}, and later supervised neural models \cite{martin2020controllable, sheang2021controllable}. 
More recently, large language models (LLMs) have enabled instruction-based simplification without task-specific fine-tuning, providing a flexible alternative to supervised approaches \cite{brown2020language, liu2023pre, cohen2026simplify}. However, prompt-based simplification remain highly sensitive to prompt formulation and often produce inconsistent simplifications, motivating research on improving the controllability of LLM-based simplification \cite{cohen2026simplify, maddela2023lens, alfear2024meta}.

RAG improves generation by incorporating relevant external information during inference \cite{lewis2020retrieval}. It has been successfully applied to a wide range of generation tasks, including question answering, summarization, and other knowledge-intensive applications \cite{parvez2021retrieval, zhao2026retrieval, zhao2025rag, cai2025comprehensive}. In these settings, retrieved documents primarily provide factual knowledge that complements the model's internal knowledge.

Unlike previous RAG applications, our work uses retrieval for example-based transformation guidance rather than factual grounding. By retrieving document-simplification examples instead of supporting evidence documents, EGP adapts retrieval to the needs of document-level simplification. 
To the best of our knowledge, this perspective has received little attention in document-level text simplification.

\section{Example-Guided Prompting}
\begin{figure*}    \centering
    \includegraphics[width=1\linewidth]{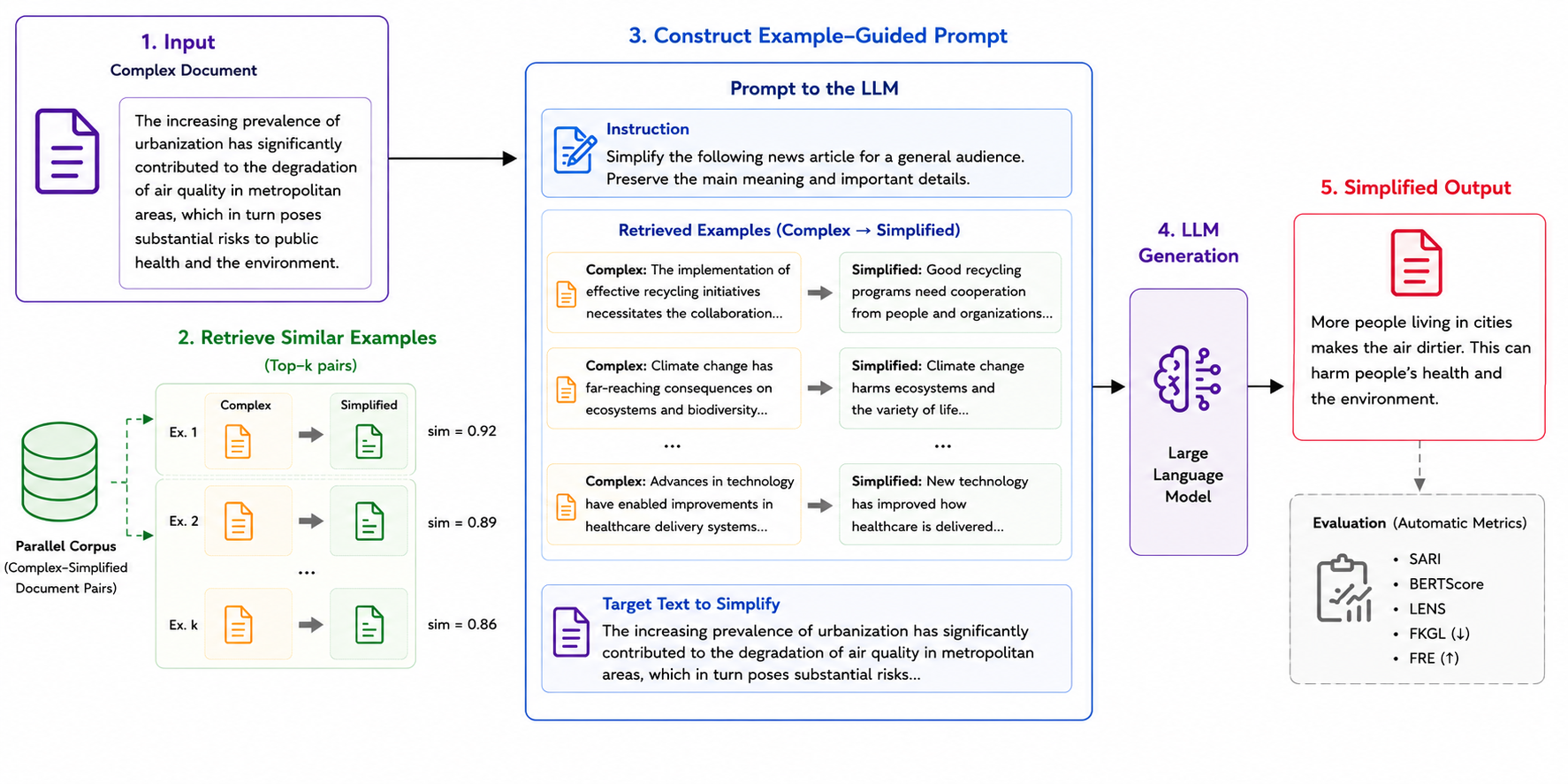}
    \caption{Overview of the proposed EGP  framework. Given a complex input document, the system retrieves semantically similar document–simplification examples from a parallel corpus and incorporates them into the prompt together with the target document. The LLM uses these examples as guidance for document-level simplification, and the generated output is evaluated using automatic measures of simplification quality, semantic preservation, and readability.}
    \label{fig:pipeline}
\end{figure*}

Figure~\ref{fig:pipeline} illustrates the EGP pipeline. Given a complex input document, EGP retrieves semantically similar document-simplification examples and incorporates them into the prompt together with the target document. The resulting prompt is passed to the LLM, which generates the simplified output. 



\section{Experiments}
\subsection{Experimental Setup}

We evaluate the proposed approach on the OneStopEnglish corpus \cite{vajjala2018onestopenglish}, using the advanced-to-elementary simplification setting.

Experiments are conducted using five state-of-the-art LLMs representing both commercial and open-weight models: GPT-4o-mini, Claude Haiku 4.5, Gemini 2.0 Flash, Sonar, and Llama 3.2.

As a preliminary experiment, we compare three prompting strategies (zero-shot, structured, and constraint-based) and select the best-performing prompt for all subsequent experiments. We then compare prompt-only generation with EGP.

For EGP, we retrieve the top-$k$ ($k=3$) most similar document--simplification examples from a parallel simplification corpus using dense retrieval. 
The retrieval component uses dense embeddings and cosine similarity to identify semantically similar documents. 
The evaluated document is excluded from the retrieval pool to prevent data leakage. 

We evaluate the generated simplifications using complementary automatic metrics that capture the principal objectives of document-level text simplification. SARI \cite{xu2016optimizing} measures simplification quality, BERTScore \cite{zhang2019bertscore} and LENS \cite{maddela2023lens} assess semantic preservation, and FKGL \cite{kincaid1975} together with FRE \cite{flesch1948new} quantify readability. This combination enables us to analyze the trade-off between producing simpler text and preserving the original meaning.

We additionally compare our approach with representative supervised (T5\footnote{T5-large-text-simplification}) and planning-based (PlanSimp) simplification systems \cite{raffel2020exploring,cripwell2023document}.
\subsection{Prompt Selection}
As shown in Appendix Table A1, constraint-based prompting consistently achieves the highest SARI while maintaining competitive semantic preservation and readability across all evaluated models. Consequently, we adopt this prompting strategy as the prompt-only baseline in subsequent experiments. 
\begin{table*}[h!]
\centering
\small
\caption{Comparison between the prompt-only baseline and EGP. Each cell reports the EGP result, with the change relative to the prompt-only baseline shown in parentheses. Positive values indicate improvement for SARI, BERTScore, LENS, and FRE, while negative values indicate improvement for FKGL.}
\label{tab:main_results}
\begin{tabular}{lccccc}
\toprule
\textbf{Model} &
\textbf{SARI} &
\textbf{BERTScore} &
\textbf{LENS} &
\textbf{FKGL$\downarrow$} &
\textbf{FRE$\uparrow$} \\\hline
\midrule
T5
&30.0  
&0.86
&46.89
&1.33 
&-3.11 \\

PlanSimp
&42.02  
&0.89 
&48.63
&-4.66 
& 16.96 \\\hline

GPT-4o-mini
& 43.78 (+0.39)
& 0.89 (+0.00)
& \textbf{72.38} (+1.28)
& -4.28 (+0.34)
& 15.85 (-1.12) \\

Gemini 2.0 Flash
& 42.51 (+3.89)
& 0.89 (+0.02)
& 64.93 (+4.20)
& -7.83 (+0.55)
& \textbf{31.69} (-2.20) \\

Sonar
& 42.29 (+1.36)
& 0.89 (+0.02)
& 49.65 (+10.09)
& -6.23 (+0.62)
& 21.33 (-3.15) \\

Claude Haiku 4.5
& \textbf{44.84} (\textbf{+4.06})
& \textbf{0.90} (+0.02)
& 70.72 (\textbf{+11.62})
& -5.13 (+1.67)
& 16.75 (-8.84) \\

Llama 3.2
& 28.53 (-5.78)
& 0.79 (-0.07)
& 35.92 (-36.72)
& 2.38 (+6.37)
& -15.93 (-30.95) \\\hline

\bottomrule
\end{tabular}
\end{table*}


\subsection{Effect of EGP}
First, we examine whether EGP improves prompt-only generation. 
Table~\ref{tab:main_results} compares prompt-only generation with the proposed EGP approach. Incorporating retrieved document-simplification examples consistently improves simplification quality for four of the five evaluated LLMs. Claude Haiku 4.5 achieves the largest improvement, increasing the SARI score from 40.78 to 44.84 while simultaneously improving semantic preservation measured by BERTScore and LENS. Similar improvements are observed for Gemini 2.0 Flash and Sonar, whereas GPT-4o-mini shows a smaller but consistent gain.

Second, we compare EGP against representative supervised and planning-based simplification systems. 
EGP consistently achieves competitive or superior performance. Claude Haiku 4.5 + EGP obtains the highest SARI score (44.84), exceeding both PlanSimp (42.02) and T5 (30.0). Gemini 2.0 Flash also outperforms PlanSimp, while Sonar reaches comparable performance without supervised training or explicit planning. 
The superiority of EGP is even more evident in semantic preservation. GPT-4o-mini and Claude Haiku 4.5 substantially outperform both T5 and PlanSimp according to LENS, improving over the planning-based baseline by more than 20 points. These results indicate that retrieved examples help preserve document meaning considerably better than existing supervised or planning-based simplification approaches. 
These results are particularly notable because both T5 and PlanSimp are task-specific document simplification systems, whereas EGP relies solely on inference-time prompting with retrieved examples. 


Finally, we analyze when EGP is most effective. 
The impact of retrieved examples is model-dependent. While Claude Haiku 4.5, Gemini 2.0 Flash, and Sonar effectively exploit retrieved simplification examples, Llama~3.2 experiences a noticeable performance degradation. This suggests that successful EGP depends not only on retrieval quality but also on an LLM's ability to integrate contextual examples during generation.

W also observe a trade-off between simplification and semantic preservation. EGP consistently improves semantic-oriented metrics, particularly BERTScore and LENS, but does not always produce the largest readability gains measured by FKGL and FRE. Retrieved examples therefore appear to encourage simplifications that remain closer to the source document while preserving more of its meaning, highlighting the need to balance readability and semantic fidelity in document-level simplification. 

\section{Discussion}
Our results indicate that EGP consistently improves the  quality of prompt-based simplification, without requiring task-specific fine-tuning. 

An important finding of this study is that the effectiveness of EGP is highly model-dependent. While Claude Haiku 4.5, Gemini 2.0 Flash, Sonar, and GPT-4o-mini successfully exploit retrieved examples, Llama 3.2 exhibits a substantial performance degradation across all evaluation metrics. This observation suggests that retrieval quality alone is insufficient; successful example-guided generation also depends on an LLM's ability to integrate retrieved examples into the generation process. Understanding which architectural or training characteristics enable this capability remains an important direction for future research.

Our experiments also reveal an interesting trade-off between readability and semantic preservation. Although EGP consistently improves SARI, BERTScore, and LENS, it generally produces smaller readability gains measured by FKGL and FRE. This suggests that retrieved examples encourage the model to generate more conservative simplifications that remain closer to the source document. Rather than aggressively simplifying the input, the model appears to prioritize preserving meaning and discourse structure while applying transformations that resemble those observed in retrieved examples. Consequently, EGP shifts the balance toward higher semantic fidelity at the cost of slightly smaller readability improvements.

Our findings suggest that retrieved examples can serve not only as a source of external knowledge, as in conventional RAG, but also as a mechanism for guiding complex text generation tasks. This perspective may extend beyond text simplification to other controlled generation problems, where examples illustrate desirable transformations rather than provide factual information. 

\section{Conclusion}
We presented EGP, an inference-time framework for document-level text simplification based on retrieved document-simplification examples. Experiments demonstrate that EGP consistently improves prompt-only generation while achieving competitive or superior performance compared with representative supervised and planning-based baselines. More broadly, our findings suggest that retrieved examples provide an effective mechanism for controlling text generation through transformation guidance rather than factual grounding. 

\section*{Limitations}
This work has several limitations that should be considered when interpreting the results.

First, the experiments were conducted using a single benchmark dataset, OneStopEnglish \cite{vajjala2018onestopenglish}, which contains educational news articles written in English. Although this corpus is widely used for evaluating document-level text simplification, the observed improvements may not directly generalize to other domains, languages, or writing styles.

Second, our evaluation relies primarily on automatic metrics. While SARI, BERTScore, LENS, FKGL, and FRE capture complementary aspects of simplification quality, they cannot fully reflect human judgments of readability, coherence, factual consistency, or usefulness for different target audiences. Human evaluation is therefore necessary to better understand how retrieved examples influence the overall quality and usability of generated simplifications.

Third, the proposed approach retrieves examples from a fixed parallel simplification corpus using a single retrieval configuration. Different retrieval strategies, retrieval corpus sizes, similarity measures, or example selection mechanisms may substantially influence the effectiveness of EGP. In particular, the relationship between retrieval quality and generation quality deserves further investigation.

Finally, the effectiveness of EGP was shown to vary considerably across LLMs. While several instruction-following models consistently benefited from retrieved examples, one model exhibited substantial performance degradation. Understanding why some models integrate retrieved examples more effectively than others remains an open research question and may provide valuable insights into the interaction between retrieval and large language model generation.

Although the proposed approach focuses on document-level text simplification, the idea of guiding generation through retrieved examples may also be applicable to other controlled text generation tasks. Evaluating the generality of this paradigm across different applications represents an important direction for future work.

\bibliography{bibliography}

\appendix

\section{Example Appendix}
\label{sec:appendix}
\begin{table*}[t]
\centering
\scriptsize
\caption{Results for different prompt strategies.}
\label{tab:prompt_only_results}
\resizebox{\textwidth}{!}{
\begin{tabular}{llrrrrr}
\hline
Model & Strategy & SARI & BERTScore &  FKGL$_{\Delta}$ & FRE$_{\Delta}$ & LENS \\
\hline
\multirow {3}{*}{Llama 3.2} & Zero-shot & 34.04 & 0.85 &   -3.41 & 14.59 & \cellcolor{gray!20}73.91 \\
 & Structured & 34.13 & 0.85   & -3.02 & 11.26 & 64.72 \\
 & Constraint & \cellcolor{gray!20}34.20 & 0.85   & \cellcolor{gray!20}-4.19 & \cellcolor{gray!20}16.75 & 69.45 \\
 \hline
\multirow {3}{*}{Gemini 2.0 Flash} & Zero-shot & 33.82 & 0.85 &  -7.42 & 31.08 & \cellcolor{gray!20}67.58 \\
 & Structured & 34.01 & 0.85   & -5.76 & 23.71 & 64.22 \\
 & Constraint & \cellcolor{gray!20}38.14 & \cellcolor{gray!20}0.87  & \cellcolor{gray!20}\textbf{-8.52} & \cellcolor{gray!20}\textbf{34.78} & 57.53 \\
 \hline
 \multirow {3}{*}{GPT-4o-mini} & Zero-shot & 39.66 & 0.88   & -3.80 & 15.00 & \cellcolor{gray!20}74.65 \\
  & Structured & 37.47 & 0.87   & -3.73 & 14.08 & 72.34 \\
 & Constraint & \cellcolor{gray!20}\textbf{43.52} & \cellcolor{gray!20}\textbf{0.89}   & \cellcolor{gray!20}-4.93 & \cellcolor{gray!20}18.16 & 69.07 \\
 \hline
 \multirow {3}{*}{Sonar} & Zero-shot & 37.45 & \cellcolor{gray!20}0.87   & \cellcolor{gray!20}-6.81 & \cellcolor{gray!20}26.88 & \cellcolor{gray!20}54.90 \\
 & Structured & 35.44 & 0.85  & -6.69 & 25.95 & 52.55 \\
 & Constraint & \cellcolor{gray!20}39.89 & 0.86 &  -6.80 & 23.46 & 31.25 \\
 \hline
\multirow {3}{*}{Claude Haiku 4.5} & Zero-shot & 35.45 & 0.86 &  -5.30 & 19.73 & \cellcolor{gray!20}55.23 \\
 & Structured & 36.26 & 0.86   & -5.13 & 17.96 & 51.71 \\
 & Constraint & \cellcolor{gray!20}39.78 & \cellcolor{gray!20}0.87 &  \cellcolor{gray!20}-6.58 & \cellcolor{gray!20}23.53 & 53.67 \\
\hline
\end{tabular}
}
\end{table*}
\subsection{Prompts}

\subsubsection{Zero-shot Prompt Template}


\begin{quote}
You are given a news article written at an advanced reading level.

Your task is to rewrite the text in simpler language for an elementary-level reader.

Preserve the original meaning and main ideas.

Use clear and short sentences.

\textbf{Text:}

\texttt{\{INPUT\_TEXT\}}
\end{quote}

\subsubsection{Structured Prompt Template}


\begin{quote}
You are given a news article written at an advanced reading level.

Your task is to rewrite the text for an elementary-level reader by following these steps:

\begin{enumerate}
    \item Identify the main ideas in the text.
    \item Remove or shorten secondary details that are not essential.
    \item Replace complex or technical words with simpler alternatives.
    \item Split long or complex sentences into shorter ones.
    \item Verify that all original facts and main ideas are preserved.
\end{enumerate}

\textbf{Text:}

\texttt{\{INPUT\_TEXT\}}
\end{quote}

\subsubsection{Constraint Prompt Template}


\begin{quote}
You are given a news article written at an advanced reading level.

Your task is to rewrite the text for an elementary-level reader under the following constraints:

\begin{itemize}
    \item Do not add, remove, or change factual information.
    \item Keep all names, places, organizations, and numbers exactly as in the original text.
    \item Use common, high-frequency words suitable for an elementary-level reader.
    \item Limit sentence length to a maximum of 12--15 words.
    \item Prefer simple and direct sentence structures.
\end{itemize}

\textbf{Text:}

\texttt{\{INPUT\_TEXT\}}
\end{quote}

\end{document}